\documentclass[a4paper,conference]{IEEEtran}
\usepackage[pdftex]{graphicx}
\usepackage[none]{hyphenat}   
\usepackage{epstopdf}
\usepackage{makecell}
\usepackage{adjustbox}
\usepackage{algorithm}
\usepackage{algorithmic}
\usepackage{multirow}
\usepackage{array}
\usepackage{amsfonts}
\usepackage{mathtools}
\usepackage{subfig}
\usepackage{amsmath}
\usepackage{url}

\ifCLASSINFOpdf
\else
\fi
\begin{document}
%
\title{Lidar-Camera Co-Training \\for Semi-Supervised Road Detection}

\author{\IEEEauthorblockN{Luca Caltagirone, Lennart Svensson, and Mattias Wahde}
\IEEEauthorblockA{Chalmers University of Technology\\ Gothenburg, Sweden\\ \{name.surname\}@chalmers.se}
\and
\IEEEauthorblockN{Martin Sanfridson}
\IEEEauthorblockA{AB Volvo, GTT, Vehicle Automation\\ Gothenburg, Sweden \\ martin.sanfridson@volvo.com}}


%


\maketitle

\begin{abstract}
Recent advances in the field of machine learning and computer vision have enabled the development of fast and accurate road detectors. Commonly such systems are trained within a supervised learning paradigm where both an input sensor's data and the corresponding ground truth label must be provided. The task of generating labels is commonly carried out by human annotators and it is notoriously time consuming and expensive. In this work, it is shown that a semi-supervised approach known as co-training can provide significant F1-score average improvements compared to supervised learning. In co-training, two classifiers acting on different views of the data cooperatively improve each other's performance by leveraging unlabeled examples. Depending on the amount of labeled data used, the improvements ranged from 1.12 to 6.10 percentage points for a camera-based road detector and from 1.04 to 8.14 percentage points for a lidar-based road detector. Lastly, the co-training algorithm is validated on the KITTI road benchmark, achieving high performance using only 36 labeled training examples together with several  thousands unlabeled ones.
\end{abstract}


%
\IEEEpeerreviewmaketitle

\section{Introduction}
An important step towards the realization of fully autonomous road vehicles is the development of robust and accurate road detectors. Once the road is correctly identified, it is possible to plan where to go and how to get there. This problem (i.e., free road surface detection) has been studied for decades and many approaches have been developed.
Until recently, most methods relied on hand-crafted features (e.g., edge detectors, color appearance, local shape descriptors)~\cite{HillelEtaAl2014} that often provide poor generalization because they are based on assumptions about the road geometry or appearance that might be valid in some settings but not in others.

A more promising strategy is to use approaches such as deep neural networks (DNNs) that can automatically learn from the data the features that are most important for solving a given task. For example, the authors of \cite{CaltagironeEtAl2017} proposed a fully convolutional neural network (FCN) to carry out road segmentation, by fusing lidar and camera data, that achieved state-of-the-art performance. 
The dominant paradigm for training DNNs, like the FCN mentioned above, is supervised learning,  which relies on the availability of ground truth labels, such as the class describing the content of an image, or the segmentation mask defining the boundaries of the road. Supervised learning has shown to be very effective; however, the annotation process needed for generating the labels is time consuming and expensive. 

Considering that unlabeled data is often much easier to obtain, semi-supervised learning~\cite{zhu2005semi} approaches have been developed that can exploit it, together with labeled data, in order to improve classification accuracy.
By learning from unlabeled data, smaller labeled data sets are needed in order to achieve comparable performance to a system trained only in a supervised fashion. For example, Laine \textit{et al.}~\cite{Laine2017} set new records on two popular semi-supervised learning benchmarks by using an approach called temporal ensembling, where multiple predictions, obtained at different time steps, are aggregated and then used as labels for computing an unsupervised loss.
Recently, Yalniz \textit{et al.}~\cite{yalniz2019billion} showed that semi-supervised learning can also boost performance when large labeled training sets are available.  By leveraging billions of unlabeled images, their approach was able to improve the accuracy of image classifiers pretrained with millions of labeled examples. That work relied on the teacher-student paradigm, where a teacher network generates labels for unlabeled data that are afterwards used for training a student network.

This paper considers the problem of road detection in lidar point clouds and camera images, and makes the following two contributions. First, a semi-supervised learning algorithm, inspired by the teacher-student paradigm, is proposed in order to leverage the complementarity of the two sensing modalities for improving performance. This approach is a variant of the co-training algorithm~\cite{BlumEtAl1998}, where the teacher-student roles are alternated between the lidar- and the camera-based road detectors. Second, it is shown that the co-training algorithm can provide significant  performance improvements by leveraging unlabeled data, especially in settings where few labeled examples are available. This is demonstrated by carrying out an extensive study that considers multiple sizes of labeled data sets. The proposed approach is further validated on a well-known benchmark. To the best of our knowledge, this work presents the first application of the co-training algorithm to the problem of road detection.

The paper is organized as follows: In Section~\ref{sect:cotrain}, the original co-training algorithm and the modified version proposed in this work are described in detail. The data sets and data preprocessing are discussed in Section~\ref{sect:dataset}. The experiments and discussion of the results are provided in Section~\ref{sect:experimental_results}. Lastly, the conclusions and future work are presented in Section~\ref{sect:conclusions}. 

\section{Co-training}
\label{sect:cotrain}
\begin{algorithm}[t]
	\caption{Co-Training~\cite{BlumEtAl1998}}
	\label{Algorithm:CotrainBlum}
	\begin{algorithmic} 
		\STATE \textbf{Require:} A two-view set $\mathcal{T}$ of labeled examples
		\STATE \textbf{Require:} A two-view set $\mathcal{U}$ of unlabeled examples 
		\STATE \textbf{Require:} Two classifiers $f_{1}$ and $f_{2}$
		\STATE Sample random subset $\mathcal{U}'\subset \mathcal{U}$
		\FORALL{\texttt{\detokenize{iteration}}}
		\STATE Train classifier $f_{1}$ on view 1 of $\mathcal{T}$
		\STATE Train classifier $f_{2}$ on view 2 of $\mathcal{T}$
		\FORALL{example $(x_{1}, x_{2})\in\mathcal{U}'$}
		\STATE Compute prediction $\tilde{y_{1}}=f_{1}(x_{1})$ 
		\STATE Compute prediction $\tilde{y_{2}}=f_{2}(x_{2})$ 
		\ENDFOR		
		\STATE 		Select most confident $p$ positive examples and $n$	negative examples classified by $f_{1}$
		\STATE 		Select most confident $p$ positive examples and $n$	negative examples classified by $f_{2}$
		\STATE 		Add the selected $2p+2n$ self-labeled examples to $\mathcal{T}$
		\STATE		Add $2p+2n$ randomly selected examples from $\mathcal{U}$ to $\mathcal{U}'$
		\ENDFOR
	\end{algorithmic}
\end{algorithm}
Co-training is a semi-supervised learning algorithm that can be applied to problems where the \textit{instance space} is partitionable into two independent \textit{views}. The instance space is an abstraction of the input space (e.g., a road scene) associated with a classification problem, whereas the views contain the actual data (e.g., point clouds, color images, etc.) that will be consumed by the classifiers.
Within this framework, the predictions obtained in one view are used as labels in the other view and can be leveraged for boosting performance. 
For example, the predictions of a lidar-based road detector, which is generally unaffected by the environment illumination, could be exploited by a camera-based detector in order to learn more discriminative features in challenging lighting conditions.  

\subsection{Original co-training algorithm}
The co-training algorithm was first introduced by Blum and Mitchell~\cite{BlumEtAl1998} who applied it to the problem of web-page classification. In this approach, it is assumed that the instance space can be partitioned into two views that are individually sufficient for a correct classification and conditionally independent given the label. Under these assumptions, Dasgupta \textit{et al.}~\cite{DasguptaEtAl2002} proved 
that by minimizing the disagreement on unlabeled data, the classifiers' respective generalization errors are also decreased. This result formally describes the \textit{consensus principle}, one of two principles that ensures the success of co-training and more generally multi-view learning~\cite{XuEtAl2013}. The other principle is known as the \textit{complementary principle}, which states that each view might contain useful information not available to the other. During co-training, this knowledge is exchanged through the unlabeled examples and can lead to better generalization.

It is worthwhile to point out that the assumption of conditional independence of the two views given the label is very strong and rarely satisfied in real problems. In this regard, Abney~\cite{Abney2002} demonstrated that a weak rule dependence is sufficient for the success of co-training. Balcan \textit{et al.}~\cite{balcan2005co} proposed a weaker expansion assumption on the underlying data distribution and proved that it is also sufficient for iterative co-training to succeed. When two independent views of the instance space are not available, it is still possible to use variants of the co-training algorithm that instead rely on classifiers with different inductive biases to bootstrap learning. For example, Wang \textit{et al.}~\cite{WangEtAl2007} presented a probably-approximately-correct (PAC) analysis showing that co-training can improve generalization as long as the two classifiers have large difference. 

The original co-training algorithm~\cite{BlumEtAl1998} is presented in Algorithm~\ref{Algorithm:CotrainBlum}: Given a labeled set $\mathcal{T}$ and an unlabeled set $\mathcal{U}$, two classifiers $f_{1}$ and $f_{2}$ are first trained in a supervised fashion using examples drawn from $\mathcal{T}$. Afterwards, each classifier is used to predict the labels of examples drawn from a random subset $\mathcal{U'}$ of $\mathcal{U}$. The most confident predictions are then added to the labeled set $\mathcal{T}$. The two steps of supervised training and unsupervised labeling are then repeated for a fixed number of iterations.

\subsection{Modified co-training algorithm}
\label{sect:deep_cotrain}
Deep neural networks commonly require long training time and heavy computational resources, in particular when working with high resolution inputs such as images and considering tasks like semantic segmentation. The original formulation of the co-training algorithm requires re-training the classifiers from scratch several times which would be too time-consuming in our case. 

For this reason, the algorithm has been modified by having one supervised learning phase followed by a semi-supervised phase where the two classifiers alternatively assume the role of teacher and student. The semi-supervised phase is a training loop consisting of two steps. The first step involves a supervised update of the student with a labeled example. In the second step, an unlabeled example drawn from $\mathcal{U}$ is shown to both the teacher and the student classifiers.  The teacher's prediction is then used as label for the student. In this formulation of the co-training algorithm, the labeled training set $\mathcal{T}$ is therefore not expanded with the labels generated by the classifiers. Considering that their predictions become increasingly more accurate over time, it is more appropriate to regenerate the labels instead of fixing them once and for all. The modified version of the co-training algorithm is presented in Algorithm~\ref{Algorithm:CotrainLuca}.

\begin{algorithm}[t!]
	\caption{Modified Co-Training}
	\label{Algorithm:CotrainLuca}
	\begin{algorithmic} 
		\STATE \textbf{Require:} A two-view set $\mathcal{T}$ of labeled examples
		\STATE \textbf{Require:} A two-view set $\mathcal{U}$ of unlabeled examples 
		\STATE \textbf{Require:} A CNN classifier $f_{1}$ with parameters $\theta_{1}$.
		\STATE \textbf{Require:} A CNN classifier $f_{2}$ with parameters $\theta_{2}$.
		\STATE Train classifier $f_{1}$ on view 1 of $\mathcal{T}$ 
		\STATE Train classifier $f_{2}$ on view 2 of $\mathcal{T}$ 		
		\STATE Set $f_{1}$ as student, $f_{s}$, and $f_{2}$ as teacher, $f_{t}$
		\FORALL{\texttt{\detokenize{iteration}}}
		\STATE 		Get example $(x_{s},x_{t},y)\in\mathcal{T}$ and compute the student prediction $\tilde{y_{s}}=f_{s}(x_s;\theta_{s})$ 
		\STATE Update $\theta_{s}$ using the cross-entropy loss and label $y$ 
		\STATE 		Get example $(x_{s},x_{t})\in\mathcal{U}$. Compute the teacher prediction $\tilde{y_{t}}=f_{t}(x_t;\theta_{t})$ and the student prediction $\tilde{y_{s}}=f_{s}(x_s;\theta_{s})$ 
		\STATE	Update $\theta_{s}$ using the KL-divergence loss weighted by factor  $\lambda_{\rm cot}$ and by considering $\tilde{y_{t}}$ as the target distribution 	\nolinebreak
		\STATE	 Swap student-teacher roles
		\ENDFOR
	\end{algorithmic}
\end{algorithm}

\subsection{Loss function}
The co-training algorithm tries to accomplish two different objectives: (1) error minimization on the manually labeled examples, and (2) agreement maximization on the unlabeled examples. Let us consider a teacher classifier $f_{t}$ and a student classifier $f_{s}$, parametrized by weights $\theta_{t}$ and $\theta_{s}$, respectively. Let us also consider a labeled example $(x_{s},x_{t},y)\in\mathcal{T}$, where $x_{s}$ is the student's view of the input, and $x_{t}$ is the  teacher's view. The supervised loss is then given by the cross entropy, denoted as $H$, between the student's prediction $f_{s}(x_{s}; \theta_{s})$ and the label $y$, that is:
\begin{equation}
\mathcal{L}_{\text{sup}} = \mathbb{E}_{(x_{s},y)\in\mathcal{T}}[H(f_{s}(x_{s}; \theta_{s}),y)].
\end{equation}

Let us now consider an unlabeled example $(x_{s},x_{t})\in\mathcal{U}$.
Both the student's prediction $f_{s}(x_{s}; \theta_{s})$  and the teacher's prediction $f_{t}(x_{t}; \theta_{t})$ represent probability distributions over the possible classes. Here, the teacher's prediction is considered as the true distribution whereas the student's prediction is regarded as an approximation. The maximization of their agreement can be then achieved by bringing the student's prediction closer to the teacher's. By considering that the Kullback-Leibler divergence, denoted as $D_{\text{KL}}$, is a natural measure of difference between probability distributions, the co-training loss is implemented as follows:
\begin{equation}
\mathcal{L}_{\text{cot}} =\mathbb{E}_{(x_{s},x_{t})\in \mathcal{U}}[D_{\text{KL}}(f_{t}(x_{t}; \theta_{t})||f_{s}(x_{s}; \theta_{s}))].
\end{equation}
It should be noted that teacher-student roles are switched at the end of each iteration, so it is not assumed that one view is better than the other. Compared to the original co-training algorithm (see Algorithm~\ref{Algorithm:CotrainBlum}), this approach considers soft-labels (continuous values) instead of hard-labels (discrete values) for the target distribution. This makes the formulation of the loss function simpler by removing the need of manually setting classification thresholds.
The total loss is then given by the weighted sum:
\begin{equation}
\label{Equation:loss}
\mathcal{L} = \mathcal{L}_{\text{sup}} + \lambda_{\text{cot}}\mathcal{L}_{\text{cot}},
\end{equation}
where the hyperparameter $\lambda_{\text{cot}}$ is a weight factor that must be tuned according to the problem at hand.

\subsection{Related work}
Recently, Qiao \textit{et al.}~\cite{qiao2018deep} trained deep neural networks for image classification using the co-training algorithm. Given that two independent views were not available, the authors instead generated adversarial examples to enforce diversity of the two classifiers.  Peng \textit{et al.}~\cite{peng2019deep} extended the previous work in order to carry out semantic segmentation of medical images.
The co-training algorithm was also used in \cite{cheng2014semi} for the task of object classification in RGB-D images acquired with a Kinect sensor. Color and depth features were learned using unsupervised learning and classification was carried out using linear SVMs. Lastly, Han \textit{et al.}~\cite{Han2018} proposed weakly supervised and semi-supervised training procedures based on generative adversarial networks to boost the performance of camera-based road detectors.

\section{Data set}
\label{sect:dataset}
This work makes use of two data sets, the KITTI raw data set \cite{GeigerEtAl2013} and the KITTI road data set \cite{FritschEtAl2013}. 
The raw data set consists of many driving sequences recorded over several days in urban, rural, and highway roads in daytime and fair weather conditions. The sensor setup used for recording the sequences included four cameras and a high-resolution lidar Velodyne HDL-64E. In this work, only one of the color cameras and the lidar are considered. Therefore, each example in a driving sequence consists of an RGB image and a point cloud. It should be noted that the road ground truth is not available for most of the examples in the raw sequences.  In this work, only a subset of the raw data set is considered which includes 59 sequences ranging in length from few seconds to few minutes, for a total of 41\,896 examples

The road data set contains 289 labeled examples. As for the previous data set, each example consists of a color image and a point cloud; however, in this case the road ground truth is also available. Figure~\ref{fig:example_labeled} shows an example of a road scene and its corresponding road label. The road data set is split into three balanced broad categories, urban marked (UM), urban multiple marked (UMM), and urban unmarked (UU) according to the presence and number of marked lanes or lack thereof. 
It is important to remark that the road data set is a subset of the raw data set, that is, each labeled example can be mapped to a specific frame in a driving sequence. This makes possible (see Sect.~\ref{sect:data_set_splits}) to find all the frames that are temporally adjacent to the labeled example.

\subsection{Data preprocessing}
\label{subsec:datapreprocessing}
Several strategies have been developed to process point clouds with deep neural networks, see for example~\cite{CaltagironeEtAl2017} and ~\cite{ZhouEtAl2018}. In this work, the lidar point cloud is simply projected into the camera plane in order to generate a three-channel tensor with the same width and height of the RGB image, and such that each channel encodes one of the 3D spatial coordinates~\cite{CaltagironeEtAl2019}. By doing so, it is straightforward to establish a one-to-one correspondence between the color information contained in the RGB image and the spatial information in the point cloud.

\begin{figure}[t!]
	\centering
	\setkeys{Gin}{width=\columnwidth}
	\includegraphics{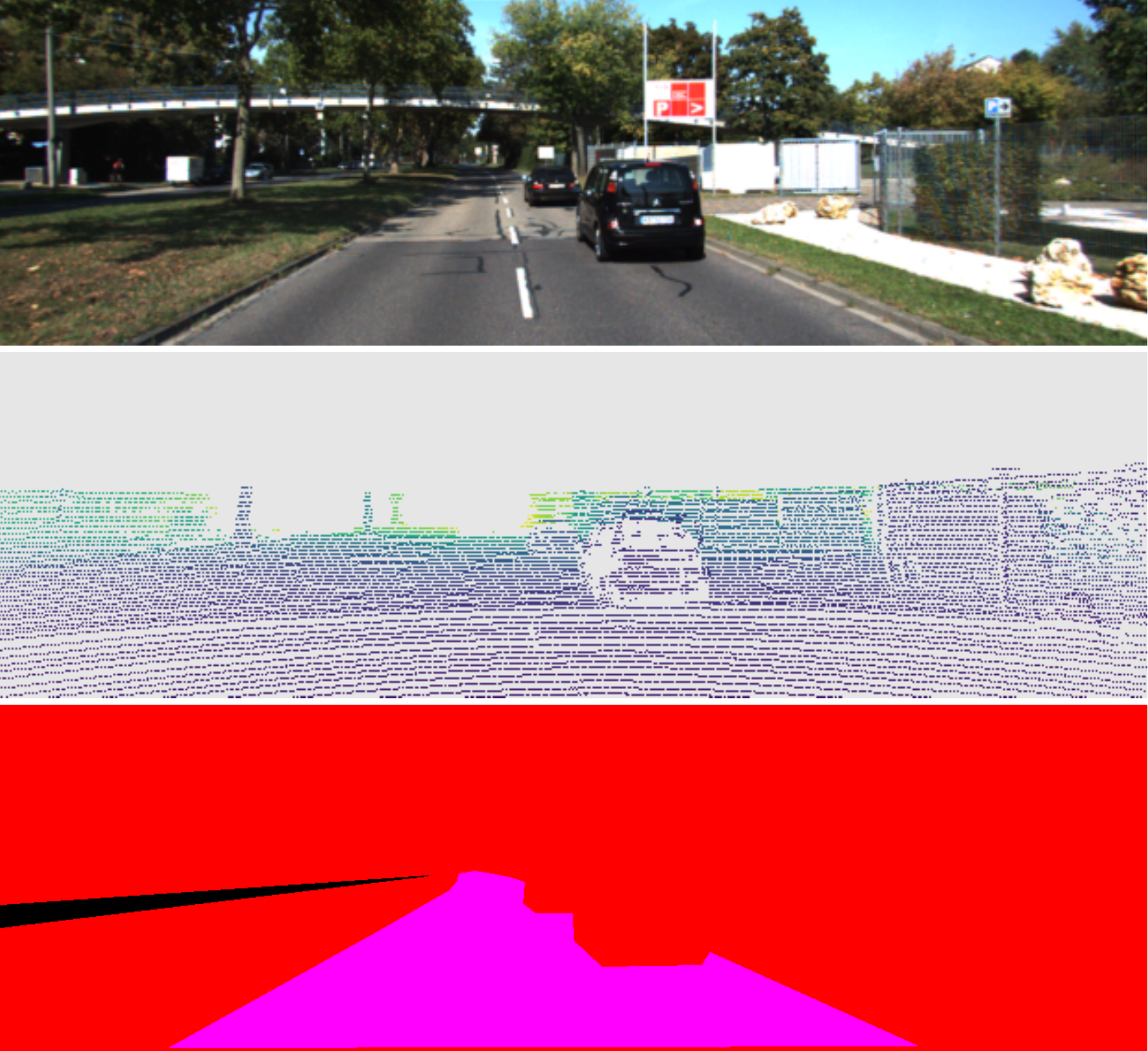}
	\caption{Labeled example from the KITTI road data set. The top panel shows the camera RGB image. The middle panel contains the corresponding Z-coordinate image (see Sect.~\ref{subsec:datapreprocessing}). 
		Lastly, the bottom panel shows the road ground truth, where the magenta pixels denote road, the red pixels are not-road, and the black pixels are ignored.}
	\label{fig:example_labeled}
\end{figure}

A point cloud acquired with a Velodyne HDL-64E consists of approximately 100k points where each point $p$ is specified by its spatial coordinates in the lidar coordinate system, that is $p = [x, y, z, 1]^{\text{T}}$. Given the lidar-camera transformation matrix $\textbf{T}$, the rectification matrix $\textbf{R}$, and the camera projection matrix $\textbf{P}$, it is possible to calculate the column position, $u$, the row position, $v$, and the scaling factor $\alpha$, where the projection of $p$ intersects the camera plane, by solving the following expression $\alpha \hspace{1mm} [u, v, 1]^{\text{T}} = \textbf{P} \hspace{1mm} \textbf{R}\hspace{1mm}\textbf{T}\hspace{1mm} p$.
This procedure is applied to every point in the point cloud, while discarding points such that $\alpha<0$ or when $[u, v]$ falls outside the image.
By using the above procedure, three images denoted as X, Y, and Z are generated where each pixel contains the $x$, $y$, and $z$ coordinates of the 3D point that was projected into it (see Fig.~\ref{fig:example_labeled} for an example of Z-coordinate image).

\section{Experiments and discussion}
\label{sect:experimental_results}
\subsection{Model and Training procedure}
The experiments reported in this work were carried out using two popular semantic segmentation networks, U-Net~\cite{RonnebergerEtAl2015} for the main study described in Sect.~\ref{sect:main_experiment} and FCN-ResNet50~\cite{he2016deep} for generating the results submitted to the KITTI road benchmark reported in Sect.~\ref{sect:kitti_benchmark}. Both neural networks were implemented and trained using the deep learning library PyTorch~\cite{paszke2017automatic}.

The initial supervised phase (see Sect.~\ref{sect:deep_cotrain}) was carried out for 65\,000 steps. Each step corresponds to one example fed to the network, so the total number of training iterations depended on the batch size. The semi-supervised phase was carried out for 200\,000 steps ($\text{batch size}=9$, $\lambda_{\text{cot}}=1$) for the main experiment, and for 300\,000 steps ($\text{batch size}=3$, $\lambda_{\text{cot}}=10$) for the KITTI benchmark. In all cases, Adam optimization~\cite{KingmaEtAl2014} was used. 
The learning rate $\eta$ was decayed using the poly learning policy \cite{OliveiraEtAl2016} implemented as $\eta(i) = \eta_{0}(1-\frac{i}{M})^{\alpha}$, 
where $i$ denotes the current step number, $\eta_{0}$ is the starting learning rate which was set to 0.0005, $M$ is the maximum number of training steps, and $\alpha$ was set to 0.9.
Data augmentation was also carried out by applying random rotations in the range $[-20^{\circ}, 20^{\circ}]$ about the center of the images and random color (brightness, contrast, saturation, and hue) jittering.
The evaluation measure used in the experiments is the F1-score, computed as $\mbox{F1}=2\cdot(\mbox{PRE}\cdot\mbox{REC})/(\mbox{PRE}+\mbox{REC})$, where PRE and REC denote precision and recall respectively.

The color images have variable sizes of approximately $1\,242\times375$ pixels. For the main experiment, in order to reduce the memory requirements, all the images were downsampled by a factor of 2 and cropped to a size of $608\times160$ pixels.
Vertical cropping preserved the bottom part of the image given that the upper part usually contains the sky, buildings, and tree tops. For the KITTI benchmark, the images were instead kept at full resolution and zero-padded to a size of $1\,248\times384$ pixels. 
Additional information about the experiments can be found online\footnote{\url{https://github.com/luca-caltagirone/cotrain}}.

\subsection{Data set splits}
\label{sect:data_set_splits}
As mentioned in Sect.~\ref{sect:dataset}, the road data set consists of 289 labeled examples whereas the raw data set contains 41\,896 unlabeled examples. Out of these two data sets, 20 random sets $\mathcal{S}_{i}= \{\mathcal{T}_{i},\mathcal{V}_{i},\mathcal{U}\}$ were generated, each containing a labeled training set $\mathcal{T}_{i}$, a labeled validation set $\mathcal{V}_{i}$, and an unlabeled set for semi-supervised learning denoted as $\mathcal{U}$. 

The labeled training and validation sets contained 144 examples each, balanced with respect to the three coarse categories UU, UM, and UMM. Additionally, in order to better assess generalization to novel scenarios, the examples in the training and validation sets were enforced to belong to different driving sequences within each category.
Each training set $\mathcal{T}_{i}$ was further split into six subsets such that $\mathcal{T}_{i}^{9} \subset \mathcal{T}_{i}^{18} \subset \mathcal{T}_{i}^{36} \subset \mathcal{T}_{i}^{72} \subset \mathcal{T}_{i}^{108} \subset \mathcal{T}_{i}^{144} = \mathcal{T}_{i}$ where the superscript indicates the subset cardinality. The examples in each subset were drawn randomly from $\mathcal{T}_{i}$ without replacement.

The unlabeled set $\mathcal{U}$ was the same for all the splits and was obtained from the raw sequences described in Sect.~\ref{sect:dataset} where all the frames temporally adjacent to the 289 labeled examples were removed. The considered time interval extended from 10 seconds before a given labeled example occurred in the corresponding driving sequence to 10 seconds after. This was done in order to avoid showing to the networks examples that are too similar to the ones found in the validation set but, at the same time, to include examples that belong to the same domain.
Lastly, the driving sequences were sampled by considering only every fifth frame, given that temporally close frames are generally quite similar to each other. At the end of this procedure, the unlabeled set $\mathcal{U}$ contained 4\,420 examples.
\subsection{Main experiment}
\label{sect:main_experiment}
The goal of the main experiment was to investigate the performance improvement that the co-training algorithm can provide and its relation to the amount of labeled data available. For this purpose, Algorithm~\ref{Algorithm:CotrainLuca} was applied to the 20 sets $\mathcal{S}_{i}$ described in the previous section. The supervised learning baseline was obtained by applying the second phase of Algorithm~\ref{Algorithm:CotrainLuca} without including the co-training loss. 

As shown in Table~\ref{table:main_results}, the co-training algorithm increased the validation average F1-score in all the considered cases. When the training sets contained only 9 labeled examples, the road detectors showed rather low performance after supervised learning. Despite their initial poor generalization, the co-training algorithm was able to boost both detectors' average F1-score, with an average improvement of 8.14 percentage points for the lidar-based detector and 6.10 percentage points for the camera-based one. As more labeled examples were included in the training sets, both the supervised baseline and the co-training algorithm performance improved; however, co-training consistently achieved higher average F1-scores.

The best performance overall was obtained when considering training sets with 144 labeled examples. In that case, the co-trained lidar-based road detector achieved 96.57\% average F1-score, with an improvement of 1.04 percentage points over the supervised baseline. As can be noticed in Table~\ref{table:main_results}, another advantage of co-training was that the F1-score standard deviation decreased in all the considered cases. This result suggests that the co-training algorithm was able to compensate for the performance gap of data set splits where the domain of the training set and the domain of the validation set were quite different from each other thus resulting in low supervised F1-scores.

\begin{table}[t!]
	\centering
	\caption{Main results showing the validation average F1-score and standard deviation obtained by applying supervised learning and co-training to the 20 splits described in Sect.~\ref{sect:data_set_splits}). The experiment considered multiple training set sizes $N\in\{9, 18, 36, 72, 108, 144\}$ while the validation sets always included 144 examples. The unlabeled set was the same for all the splits and consisted of 4\,420 examples. The numbers within parentheses denote the average improvement with respect to the supervised baseline.}
	\label{table:main_results}
	\renewcommand{\arraystretch}{1.2}
	\begin{tabular}{lccc} 
		\cline{3-4}
		&                     & Supervised baseline & Co-training\\ 
		\hline\hline
		\multirow{2}{*}{\rotatebox[origin=c]{0}{N=9}} & 
		\multirow{1}{*}{lidar}   & $84.25\pm2.86$ & $92.39\pm 1.34 \hspace{1mm}(8.14)$ \\
		& \multirow{1}{*}{camera}  &   $86.52\pm1.67$ & $92.62\pm 0.99 \hspace{1mm}(6.10)$   \\
		\hline	
		\multirow{2}{*}{\rotatebox[origin=c]{0}{N=18}} & 
		\multirow{1}{*}{lidar}   & $87.89\pm2.49$ & $93.87\pm 1.15 \hspace{1mm}(5.98)$    \\
		& \multirow{1}{*}{camera}  & $89.92\pm1.24$ & $93.86\pm 1.05 \hspace{1mm}(3.94)$\\
		\hline
		\multirow{2}{*}{\rotatebox[origin=c]{0}{N=36}} & 
		\multirow{1}{*}{lidar}   & $91.71\pm1.52$ & $95.38\pm 0.35 \hspace{1mm}(3.67)$    \\
		& \multirow{1}{*}{camera} & $92.43\pm0.63$ & $95.21\pm 0.46 \hspace{1mm}(2.78)$\\
		\hline
		\multirow{2}{*}{\rotatebox[origin=c]{0}{N=72}} & 
		\multirow{1}{*}{lidar}  & $94.30\pm0.77$ & $96.11\pm 0.34 \hspace{1mm}(1.81)$  \\	
		& \multirow{1}{*}{camera} & $93.97\pm0.61$ & $95.92\pm 0.34 \hspace{1mm}(1.95)$ \\
		\hline
		\multirow{2}{*}{\rotatebox[origin=c]{0}{N=108}} & 
		\multirow{1}{*}{lidar}  & $95.07\pm0.55$ & $96.44\pm 0.35 \hspace{1mm}(1.37)$  \\	
		& \multirow{1}{*}{camera} & $94.79\pm0.57$ & $96.23\pm 0.43 \hspace{1mm}(1.44)$  \\
		\hline
		\multirow{2}{*}{\rotatebox[origin=c]{0}{N=144}} & 
		\multirow{1}{*}{lidar}    & $95.53\pm0.49$ & $96.57\pm 0.26 \hspace{1mm}(1.04)$\\
		& \multirow{1}{*}{camera}  & $95.28\pm0.50$ & $96.40\pm 0.40 \hspace{1mm}(1.12)$ \\
		\hline	
	\end{tabular}
\end{table}

\subsection{KITTI benchmark}
\label{sect:kitti_benchmark}
The purpose of this experiment was to evaluate how well the co-trained road detectors can generalize to the unseen data of the KITTI road test set. The co-training algorithm was applied by considering 90 labeled examples randomly drawn from the KITTI road data set, where 36 were assigned to the training set and 54 to the validation set. The unlabeled data set was obtained by considering all the KITTI raw sequences sampled at 2 Hz where the frames temporally close (from 3 seconds before to 3 seconds after) to the labeled examples were removed. By following this procedure, the unlabeled data set contained 6\,445 examples.
Two FCN-Resnet50, one consuming only lidar data and the other only camera images, were first trained individually in a supervised fashion for 65\,000 steps and then co-trained for 300\,000 additional steps. 
As for the previous experiment, the baseline supervised networks were obtained by applying the second phase of Algorithm~\ref{Algorithm:CotrainLuca} without including the co-training loss. 

Only the camera-based road detectors were considered for evaluation on the KITTI road test set. The first one, RGB36-Super, was the CNN trained exclusively with labeled examples, whereas the second one, RGB36-Cotrain, was the CNN trained also with unlabeled data. The underlying motivation was to simulate the scenario where the data collection vehicle is equipped with both cameras and lidars, that can be leveraged for using co-training, but the consumer vehicle has only cameras available.  

\begin{figure}[t!]
	\centering
	\setkeys{Gin}{width=\columnwidth}
	\includegraphics{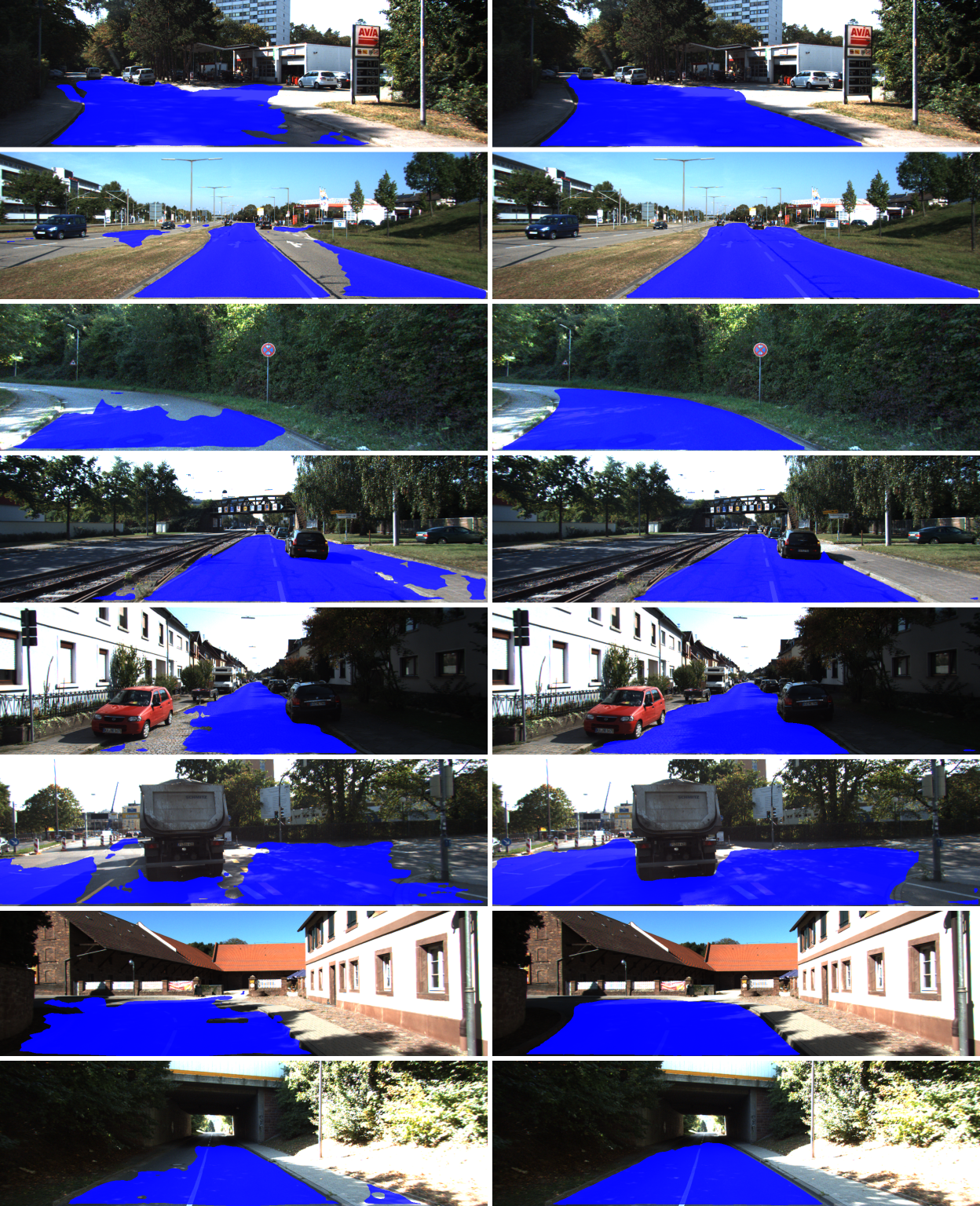}
	\caption{Road detections obtained with RGB36-Super, a CNN trained in a purely supervised fashion (left column), and with RGB36-Cotrain, a CNN trained with the co-training algorithm (right column). The reported results show that RGB36-Cotrain can better deal with scenarios  presenting challenging illumination or road pavements. }
	\label{fig:comparison_test}
\end{figure}

As shown in Table~\ref{table:kitti_results}, RGB36-Super achieved 92.94\% F1-score and ranked 15\textsuperscript{th} among published methods. RGB36-Cotrain instead reached 95.55\%  F1-score, an improvement of 2.61 percentage points compared with the supervised baseline, and ranked 4\textsuperscript{th} in the benchmark. Some qualitative results of road detections are shown in Fig.~\ref{fig:comparison_test}.
\begin{table}[t!]
	\centering
	\caption{KITTI road benchmark results (in \%) on the URBAN\_ROAD category.  Only results of published methods are reported. Training can either be \textit{super} for supervised learning, or \textit{semi} for semi-supervised learning. The numbers within parentheses denote the number of labeled training examples used. MaxF is the maximum F1-score.} 
	\label{table:kitti_results}
	\resizebox{1\columnwidth}{!}{%
		\renewcommand{\arraystretch}{1.1}
		\begin{tabular}{ccccccc}
			\hline
			{Rank} & {Method} & Sensor & Training &{MaxF} & {PRE} & {REC}\\		\hline 			\hline
			1 & PLARD \cite{Chen2019} & lid-cam & super (289) & 97.03  & 97.19 & 96.88 \\
			2 & LidCamNet \cite{CaltagironeEtAl2019} & lid-cam & super (239) &96.03  & 96.23 & 95.83\\
			4 & \textbf{RGB36-Cotrain} & cam & semi (36)& 95.55  & 95.68  & 95.42\\			
			5 & SSLGAN \cite{Han2018}& cam & semi (266) & 95.53 & 95.84 & 95.24\\
			10 & LoDNN \cite{CaltagironeEtAl2017} &  lidar & super (259) & 94.07   & 92.81  & 95.37\\	
			15 & \textbf{RGB36-Super} & cam & super (36) & 92.94  & 93.14  & 92.74\\					
			\hline
	\end{tabular}}
\end{table}

\section{Conclusions and future work}
\label{sect:conclusions}
This work introduced a variant of the co-training algorithm that is suitable for training deep neural networks.  The proposed approach was applied to the problem of road detection by considering two complementary and independent views of the environment obtained with a color camera and a high-resolution lidar, respectively. An extensive study that considered multiple sizes of labeled training sets was carried out. The results highlighted the significant performance boost that co-training can provide, with respect to supervised learning, ranging in average from 1.04 to 8.14 percentage points F1-score in the case of a lidar-based detector and from 1.12 to 6.10 percentage points for a camera-based one. The proposed approach was also evaluated on the KITTI road benchmark and ranked among the top-performing methods while using only a small amount of labeled data.

For future work, it would be of interest to evaluate the co-training algorithm by considering a larger data set with broader variability in terms of illumination, weather, and road types, and more semantic classes. It would also be of interest to extend the proposed approach to include larger ensembles of networks or to combine it with other semi-supervised and unsupervised methods.

\section*{Acknowledgment}
The authors gratefully acknowledge financial support from Vinnova/FFI, grant number 2014-06239.
This work was also partially supported by the Wallenberg Al, Autonomous Systems and Software Program (WASP) funded by the Knut and Alice Wallenberg Foundation.


\bibliographystyle{IEEEtran}
\bibliography{IEEEabrv,bibliography}

\begin{thebibliography}{10}
\providecommand{\url}[1]{#1}
\csname url@samestyle\endcsname
\providecommand{\newblock}{\relax}
\providecommand{\bibinfo}[2]{#2}
\providecommand{\BIBentrySTDinterwordspacing}{\spaceskip=0pt\relax}
\providecommand{\BIBentryALTinterwordstretchfactor}{4}
\providecommand{\BIBentryALTinterwordspacing}{\spaceskip=\fontdimen2\font plus
\BIBentryALTinterwordstretchfactor\fontdimen3\font minus
  \fontdimen4\font\relax}
\providecommand{\BIBforeignlanguage}[2]{{%
\expandafter\ifx\csname l@#1\endcsname\relax
\typeout{** WARNING: IEEEtran.bst: No hyphenation pattern has been}%
\typeout{** loaded for the language `#1'. Using the pattern for}%
\typeout{** the default language instead.}%
\else
\language=\csname l@#1\endcsname
\fi
#2}}
\providecommand{\BIBdecl}{\relax}
\BIBdecl

\bibitem{HillelEtaAl2014}
A.~B. Hillel, R.~Lerner, D.~Levi, and G.~Raz, ``Recent progress in road and
  lane detection: a survey,'' \emph{Machine vision and applications}, vol.~25,
  no.~3, pp. 727--745, 2014.

\bibitem{CaltagironeEtAl2017}
L.~{Caltagirone}, S.~{Scheidegger}, L.~{Svensson}, and M.~{Wahde}, ``Fast
  lidar-based road detection using fully convolutional neural networks,'' in
  \emph{2017 IEEE Intelligent Vehicles Symposium (IV)}, 2017, pp. 1019--1024.

\bibitem{zhu2005semi}
X.~J. Zhu, ``Semi-supervised learning literature survey,'' University of
  Wisconsin-Madison Dept. of Computer Sciences, Tech. Rep., 2005.

\bibitem{Laine2017}
S.~Laine and T.~Aila, ``Temporal ensembling for semi-supervised learning,'' in
  \emph{Proc. International Conference on Learning Representations (ICLR)},
  2017.

\bibitem{yalniz2019billion}
I.~Z. Yalniz, H.~J{\'e}gou, K.~Chen, M.~Paluri, and D.~Mahajan, ``Billion-scale
  semi-supervised learning for image classification,'' \emph{arXiv preprint
  arXiv:1905.00546}, 2019.

\bibitem{BlumEtAl1998}
A.~Blum and T.~Mitchell, ``Combining labeled and unlabeled data with
  co-training,'' in \emph{Proceedings of the eleventh annual conference on
  Computational learning theory}.\hskip 1em plus 0.5em minus 0.4em\relax
  Citeseer, 1998, pp. 92--100.

\bibitem{DasguptaEtAl2002}
S.~Dasgupta, M.~L. Littman, and D.~A. McAllester, ``Pac generalization bounds
  for co-training,'' in \emph{Advances in Neural Information Processing Systems
  14}.\hskip 1em plus 0.5em minus 0.4em\relax MIT Press, 2002, pp. 375--382.

\bibitem{XuEtAl2013}
C.~Xu, D.~Tao, and C.~Xu, ``A survey on multi-view learning,'' \emph{arXiv
  preprint arXiv:1304.5634}, 2013.

\bibitem{Abney2002}
S.~Abney, ``Bootstrapping,'' in \emph{Proceedings of the 40th Annual Meeting on
  Association for Computational Linguistics}, 2002, pp. 360--367.

\bibitem{balcan2005co}
M.-F. Balcan, A.~Blum, and K.~Yang, ``Co-training and expansion: Towards
  bridging theory and practice,'' in \emph{Advances in neural information
  processing systems}, 2005, pp. 89--96.

\bibitem{WangEtAl2007}
W.~Wang and Z.-H. Zhou, ``Analyzing co-training style algorithms,'' in
  \emph{European conference on machine learning}.\hskip 1em plus 0.5em minus
  0.4em\relax Springer, 2007, pp. 454--465.

\bibitem{qiao2018deep}
S.~Qiao, W.~Shen, Z.~Zhang, B.~Wang, and A.~Yuille, ``Deep co-training for
  semi-supervised image recognition,'' in \emph{Proceedings of the European
  Conference on Computer Vision (ECCV)}, 2018, pp. 135--152.

\bibitem{peng2019deep}
J.~Peng, G.~Estradab, M.~Pedersoli, and C.~Desrosiers, ``Deep co-training for
  semi-supervised image segmentation,'' \emph{arXiv preprint arXiv:1903.11233},
  2019.

\bibitem{cheng2014semi}
Y.~Cheng, X.~Zhao, K.~Huang, and T.~Tan, ``Semi-supervised learning for rgb-d
  object recognition,'' in \emph{2014 22nd International Conference on Pattern
  Recognition}.\hskip 1em plus 0.5em minus 0.4em\relax IEEE, 2014, pp.
  2377--2382.

\bibitem{Han2018}
X.~Han, J.~Lu, C.~Zhao, S.~You, and H.~Li, ``Semisupervised and weakly
  supervised road detection based on generative adversarial networks,''
  \emph{IEEE Signal Processing Letters}, vol.~25, no.~4, pp. 551--555, 2018.

\bibitem{GeigerEtAl2013}
A.~Geiger, P.~Lenz, C.~Stiller, and R.~Urtasun, ``Vision meets robotics: The
  kitti dataset,'' \emph{International Journal of Robotics Research}, 2013.

\bibitem{FritschEtAl2013}
J.~Fritsch, T.~Kuehnl, and A.~Geiger, ``A new performance measure and
  evaluation benchmark for road detection algorithms,'' in \emph{International
  Conference on Intelligent Transportation Systems (ITSC)}, 2013.

\bibitem{ZhouEtAl2018}
Y.~Zhou and O.~Tuzel, ``Voxelnet: End-to-end learning for point cloud based 3d
  object detection,'' in \emph{Proceedings of the IEEE Conference on Computer
  Vision and Pattern Recognition}, 2018, pp. 4490--4499.

\bibitem{CaltagironeEtAl2019}
L.~Caltagirone, M.~Bellone, L.~Svensson, and M.~Wahde, ``Lidar--camera fusion
  for road detection using fully convolutional neural networks,''
  \emph{Robotics and Autonomous Systems}, vol. 111, pp. 125--131, 2019.

\bibitem{RonnebergerEtAl2015}
O.~Ronneberger, P.~Fischer, and T.~Brox, ``U-net: Convolutional networks for
  biomedical image segmentation,'' in \emph{International Conference on Medical
  image computing and computer-assisted intervention}.\hskip 1em plus 0.5em
  minus 0.4em\relax Springer, 2015, pp. 234--241.

\bibitem{he2016deep}
K.~He, X.~Zhang, S.~Ren, and J.~Sun, ``Deep residual learning for image
  recognition,'' in \emph{Proceedings of the IEEE conference on computer vision
  and pattern recognition}, 2016, pp. 770--778.

\bibitem{paszke2017automatic}
A.~Paszke, S.~Gross, S.~Chintala, G.~Chanan, E.~Yang, Z.~DeVito, Z.~Lin,
  A.~Desmaison, L.~Antiga, and A.~Lerer, ``Automatic differentiation in
  {PyTorch},'' in \emph{NIPS Autodiff Workshop}, 2017.

\bibitem{KingmaEtAl2014}
D.~Kingma and J.~Ba, ``Adam: A method for stochastic optimization,''
  \emph{arXiv preprint arXiv:1412.6980}, 2014.

\bibitem{OliveiraEtAl2016}
G.~L. Oliveira, W.~Burgard, and T.~Brox, ``Efficient deep models for monocular
  road segmentation,'' in \emph{IEEE/RSJ International Conference on
  Intelligent Robots and Systems (IROS)}, 2016, pp. 4885--4891.

\bibitem{Chen2019}
Z.~Chen, J.~Zhang, and D.~Tao, ``Progressive lidar adaptation for road
  detection,'' \emph{IEEE/CAA Journal of Automatica Sinica}, vol.~6, no.~3, pp.
  693--702, 2019.

\end{thebibliography}

\end{document}